\documentclass{article}

 \usepackage[final]{eiml_style_2025}

\usepackage[utf8]{inputenc} 
\usepackage[T1]{fontenc}    
\usepackage{hyperref}       
\usepackage{url}            
\usepackage{booktabs}       
\usepackage{amsfonts}       
\usepackage{nicefrac}       
\usepackage{microtype}      

\setcitestyle{numbers}

\usepackage{tikz} 
\usetikzlibrary{positioning}
\usepackage{booktabs} 
\usepackage{fvextra} 
\usepackage{graphicx}
\usepackage{nicefrac}

\usepackage{pgfplots}
\pgfplotsset{compat=1.18}
\usepgfplotslibrary{groupplots}
\usepackage{subfigure}
\usepackage{framed}

\usepackage{mathtools}

\usepackage{csvsimple-l3}
\usepackage{tabularray,siunitx,xfp}
\usepackage{xcolor}
\usepackage{pgfplots}

\newcommand{\class}{c}

\newcommand{\Classes}{\mathcal{C}}

\newcommand{\prediction}{\hat{\class}}

\newcommand{\classifier}{h}

\newcommand{\feature}[1]{f_{#1}}

\newcommand{\FeatureSet}[1]{\mathcal{F}_{#1}}
\newcommand{\features}{f}

\newcommand{\FeaturesSet}{\mathcal{F}}
\newcommand{\numfeatures}{N}

\newcommand{\prob}{P}

\newcommand{\datadistr}{\prob_{\mathrm{data}}}
\newcommand{\learneddistr}{\prob_{\mathrm{classif}}}

\newcommand{\testset}{D_{\mathrm{test}}}

\newcommand{\trainset}{D_{\mathrm{train}}}

\newcommand{\maxprob}{u_{\mathrm{m}}}

\newcommand{\entropy}{u_H}
\newcommand{\aleatunc}{u_{a}}
\newcommand{\totunc}{u_{t}}
\newcommand{\epistunc}{u_{e}}

\newcommand{\globeps}{\varepsilon_{\mathrm{glob}}}
\newcommand{\loceps}{\varepsilon_{\mathrm{loc}}}

\definecolor{darkgreen}{rgb}{0.01, 0.65, 0.18}
\definecolor{rob_blue}{rgb}{0.0, 0.75, 1.0}
\definecolor{unc_yellow}{rgb}{0.95, 0.82, 0.27}
\definecolor{hybrid_green}{rgb}{0.1, 0.7, 0.2}

\pgfplotstableread{BCW_global_eps_entropies.dat}\ARC

\csvnames{myNames}{1=\dataset, 2=\unc, 3=\glob, 4=\hybridA, 5=\mixingA, 6=\local, 7=\hybridB, 8=\mixingB}

\title{Robustness and uncertainty: two complementary aspects of the reliability of the predictions of a classifier}

\author{%
  Adri\'an Detavernier\\
  Foundations Lab for imprecise probabilities\\
  Ghent University\\
  Belgium \\
  \And
  Jasper De Bock\\
  Foundations Lab for imprecise probabilities\\
  Ghent University\\
  Belgium \\
}

\begin{document}

\maketitle

\begin{abstract}
    We consider two conceptually different approaches for assessing the reliability of the individual predictions of a classifier: Robustness Quantification (RQ) and Uncertainty Quantification (UQ).
    We compare both approaches on a number of benchmark datasets and show that there is no clear winner between the two, but that they are complementary and can be combined to obtain a hybrid approach that outperforms both RQ and UQ.
    As a byproduct of our approach, for each dataset, we also obtain an assessment of the relative importance of uncertainty and robustness as sources of unreliability.
\end{abstract}

\section{Introduction}
\label{sec:intro}
Due to its vast capabilities, AI has become ubiquitous, its use cases ranging from automating simple tasks to making decisions in high-risk settings.
In some cases, especially the ones where the stakes are high, we are not only interested in the overall performance of the model, but also in the quality or, to be more precise, the reliability of each single prediction.
If your own health is at stake for instance, you care less about how well the model performs on average; you only want to know whether you can rely on the model's prediction in your particular case.
So, in an ideal world, we'd want to know for each prediction of an AI model how reliable it is.
For the least reliable predictions, a second opinion of an expert could then be asked, more data could be collected, etc.

One of the more popular applications of AI models, and the one we focus on in this paper, is classification.
In that case, the goal of the model is to predict the correct class \(\class\) of a given instance out of a set of possible classes \(\Classes\).
An instance is usually described using a number of features (\(\numfeatures\) in total). The value \(\feature{i}\) of the \(i\)-th feature takes values in a set \(\FeatureSet{i}\), which we take to be finite because we'll restrict ourselves to discrete features. We'll call the vector \(\features \coloneq (\feature{1}, \dots, \feature{\numfeatures})\) the (set of) features of said instance, which takes values in \(\FeaturesSet \coloneq \FeatureSet{i} \times \dots \times \FeatureSet{\numfeatures}\).
In practice a classifier then, given an instance (e.g. a patient), uses its features \(\features\) (e.g. the patient's medical data) to try to predict the correct class \(\class\) (e.g. the sickness of the patient).
We'll denote the class predicted by the classifier as \(\prediction\).

For each such prediction \(\prediction\) of a classifier, we can now try to assess how reliable it is.
In this work, we consider two methods for doing so, namely \emph{uncertainty quantification} \citep{hullermeier2021aleatoric,pmlr-v180-hullermeier22a,sale2024label} and \emph{robustness quantification} \citep{correia2020robustclassificationdeepgenerative,NIPS2014_09662890,detavernier2025robustness,pmlr-v62-mauá17a}.
What both approaches have in common is that they are based on the core idea that there is a lot of uncertainty involved when learning a model from data.
Uncertainty quantification tries to quantify this uncertainty, for the prediction associated with a given instance.
Robustness quantification, on the other hand, tries to quantify the amount of (epistemic) uncertainty the model could handle, while still issuing the same prediction for the given instance, regardless of how much uncertainty there actually is.
In a side by side comparison, we've recently demonstrated that robustness quantification does a better job at assessing the reliability of the predictions of a classifier than uncertainty quantification, at least for artificial data and in the presence of distribution shift or when there was a limited amount of data \citep{detavernier2025robustness}.
We now opt for a different point of view: instead of only comparing the two and studying which one is better, why not combine them? Since, on a conceptual level, robustness and uncertainty cover different aspects of reliability, it seems plausible that such a combination should lead to even better reliability assessments. In our experiments on benchmark datasets, we demonstrate that this is indeed the case.

\section{Uncertainty and robustness for probabilistic generative classifiers}
\label{sec:UqandRQ}
Formally, a classifier \(\classifier: \FeaturesSet \to \Classes\) is a function from the set of all possible sets of features to the set of possible classes.
The uncertainty and robustness metrics we consider in this work, are designed for probabilistic classifiers and, in the case of the robustness metrics, generative ones. Since our features are discrete, such a probabilistic generative classifier is completely determined by a probability mass function \(\learneddistr\) on \(\Classes\times\FeaturesSet\).
For a given set of features \(\features\), the predicted class \(\prediction\) is then the one with the highest probability given the features:
\begin{equation*}
	\prediction \coloneq \classifier(\features) = \arg \max_{\class \in \Classes} \learneddistr(\class \vert \features),
\end{equation*}
where \(\learneddistr(\cdot \vert \features)\) is obtained from \(\learneddistr\) through Bayes' rule.
In our experiments further on, we make use of Naive Bayes Classifiers (NBC) \citep{naiveBayes}, which are probabilistic generative classifiers that assume the features to be independent given the class.

In practice, (the probability mass function of) a generative classifier \(\learneddistr\) is learned using a training set \(\trainset\) of correctly labeled instances, and its performance is then evaluated on a different set of instances, called the test set \(\testset\).
It is often assumed that these datasets are sampled from a common underlying distribution, whose probability mass function we will denote by \(\datadistr\)\footnote{
	Note that we assume that both \(\trainset\) and \(\testset\) are sampled from the same distribution.
	This need not be always the case though.
	The phenomenon of the training and test distribution not being equal is called \emph{distribution shift}.
	For a study of how uncertainty and robustness quantification perform (and compare) in the presence of distribution shift, we refer the interested reader to our recent work \citep{detavernier2025robustness}.
}.

The ideal classifier is the one for which \(\learneddistr = \datadistr\).
However, even in this ideal case, the accuracy of the predictions issued by this classifier will typically not be 100\%, meaning that even then there still is uncertainty present in the prediction.
This has to do with the intrinsic variability present in the task at hand: two instances with the same features could in practice have a different class, either due to inherent randomness or because not enough information is captured in the set of features to distinguish these cases.
This remaining uncertainty is completely captured by \(\datadistr(\cdot \vert \features)\), and we will refer to it as \emph{aleatoric uncertainty}.

Unfortunately, the case where the classifier perfectly learns \(\datadistr\) is unrealistic. The more realistic scenario is that the learned classifier differs from the ideal one, that is, \(\learneddistr \neq \datadistr\).
The fact that \(\learneddistr\) and \(\datadistr\) need not be the same, is a completely different type of uncertainty associated with classification, which we call \emph{epistemic uncertainty}.
Possible sources of this uncertainty are structural modelling assumptions (such as the independence assumption of an NBC) or the fact that \(\learneddistr\) is based on a finite (and hence possibly too small or unrepresentative) training set.

\subsection{Uncertainty quantification}

Uncertainty quantification tries to quantify either of these two sources of uncertainty, or both, in the form of a numerical uncertainty metric.
This task is extremely challenging, though, since we'll never know the distribution \(\datadistr\), nor whether \(\trainset\) or \(\testset\) are representative for \(\datadistr\).
Any estimate of the amount of aleatoric uncertainty therefore has to be based on the---most likely wrong--- distribution \(\learneddistr\).
Similarly, estimating the epistemic uncertainty or, in other words, the extent to which \(\learneddistr\) differs from \(\datadistr\), is obviously very hard since we don't know \(\datadistr\).

Nevertheless, many uncertainty metrics have been developed. In our experiments, we consider 5 of them.
A first intuitive uncertainty metric is one minus the probability of the predicted class according to \(\learneddistr(\cdot\vert\features)\), which we denote by \(\maxprob\).
In the ideal case where \(\learneddistr=\datadistr\), this would be equivalent to the probability of making a wrong decision.
It thus can be seen as an estimate of the aleatoric uncertainty for the prediction associated with $\features$.
A different attempt at estimating the aleatoric uncertainty of an instance with features \(\features\) makes use of the (Shannon) entropy.
This metric, denoted by \(\entropy\), is the entropy of \(\learneddistr(\cdot\vert\features)\).
The remaining uncertainty metrics combine entropy with ensemble techniques \citep{shaker2020aleatoric}.
These metrics try to estimate the total, aleatoric and epistemic uncertainty, and are denoted by \(\totunc\), \(\aleatunc\) and \(\epistunc\), respectively; the exact formulas are available in our previous work \citep{detavernier2025robustness}.
For a more in-depth overview of these and other uncertainty metrics, we refer to the work of \citet{hullermeier2021aleatoric}.

\subsection{Robustness quantification}
Robustness quantification takes a different approach by instead trying to numerically quantify how much epistemic uncertainty a model could handle before its prediction changes.
The idea of robustness quantification has its origin in the field of imprecise probability theory \citep{augustin2014introduction}.
Instead of only looking at what class is predicted by \(\learneddistr\), this approach considers neighborhoods of distributions around \(\learneddistr\).
If all distributions in such a neighborhood predict the same class as the one predicted by \(\learneddistr\), we call this prediction \emph{robust} w.r.t. said neighborhood.
If the prediction is robust, then the size of the neighborhood can be seen as a lower bound on the amount of epistemic uncertainty we can allow without changing the prediction.
By controlling the size of the neighborhood in a parametrized manner, we can increase this size until the prediction of the model is no longer robust, or thus until at least one distribution in the neighborhood predicts a different class.
The parameter value at which this happens can then be used as a robustness metric, indicating the amount of epistemic uncertainty the model could handle without changing this particular prediction.
Quantifying robustness this way has been successfully tried several times already in the literature, for different types of classifiers \citep{correia2020robustclassificationdeepgenerative,NIPS2014_09662890,detavernier2025robustness,pmlr-v62-mauá17a}.

There are of course numerous types of neighborhoods that can be considered, and therefore many different robustness metrics.
We restrict ourselves to two such families, and thus to two robustness metrics.
The first robustness metric, which we denote by \(\globeps\), can be applied to any probabilistic generative classifier; it considers neighborhoods of the learned (global) joint distribution \(\learneddistr\) obtained by \(\epsilon\)-contaminating~\citep{Huber1992} the latter.
The second robustness metric on the other hand, which we denote by \(\loceps\), is tailor-made for Naive Bayes models. This metric considers global neighborhoods of the Naive Bayes model \(\learneddistr\) that consist of Naive Bayes models only, obtained by \(\epsilon\)-contaminating the local models of \(\learneddistr\).
For more details about these two robustness metrics, including how to efficiently compute them, we refer to our recent work on this topic \citep{detavernier2025robustness}.

Finally, since `robustness' refers to many different concepts within ML, we'd like to stress the fact that robustness quantification is instance-based, meaning that it assesses the robustness of individual predictions. This sets it apart from the plethora of approaches that consider the robustness of a classifier as a whole, such as adversarial robustness \citep{bai2021recent,carlini2019evaluating}, robustness against distribution shift \citep{NEURIPS2020_d8330f85}, or robust optimization \citep{ben2009robust}.

\section{Evaluating reliability metrics}
Since uncertainty and robustness metrics both share the goal of trying to assess the reliability of the individual predictions of a classifier, it makes sense to refer to both of them as \emph{reliability metrics}.
Depending on the task at hand, such a metric can be used to either select the most reliable instances (for example to automate the decisions for those instances) or to select the least reliable ones, and hence the hardest ones to classify (for example to classify these manually, or collect more data for them).
A perfect reliability metric would thus be able to order all instances in such a way that if we'd start rejecting instances in that order, we would first reject all wrongly classified ones, and then the correct ones.
A straightforward way of evaluating the performance of a reliability metric is therefore to look at how well it is capable of ordering a set of instances such that the misclassified instances are rejected first.

\begin{figure}[!ht]
	\begin{tikzpicture}[scale=0.7]
		\begin{groupplot}[group style={group size=2 by 1}]

		\nextgroupplot[
			xlabel={Rejection rate},
			ylabel={Accuracy},
			width=14cm,
			height=6cm,
			enlarge x limits=false,
			legend pos=south east,
			legend entries={%
                ideal,
				$\globeps$,
				$\entropy$,
				hybrid,
			},
			]
			\addplot+ [no marks, very thick, color=black, dashed] table [x={rej_rate}, y={ideal}] {\ARC};
			\addplot+ [no marks, very thick, color=rob_blue] table [x={rej_rate}, y={robustness}] {\ARC};
			\addplot+ [no marks, very thick, color=unc_yellow] table [x={rej_rate}, y={uncertainty}] {\ARC};
			\addplot+ [no marks, very thick, color=hybrid_green] table [x={rej_rate}, y={hybrid}] {\ARC};

		\nextgroupplot[
			xlabel={\(\ln\globeps\)},
			ylabel={\(-\ln\entropy\)},
			width=6cm,
			height=6cm,
			ylabel near ticks,
			yticklabel pos=right,
			xshift=-.45cm,
			]
			\addplot [
				scatter,
				only marks,
				mark size=1.2pt,
				opacity=0.5,
				point meta=explicit symbolic,
				scatter/classes={
					0={red},
					1={darkgreen}
				},
			] table [x={global_eps}, y={entropies}, meta={pred}] {BCW_global_eps_entropies_pc.dat};

		\end{groupplot}
	\end{tikzpicture}

	\caption{\footnotesize ARC (left) and point cloud (right) for the Breast Cancer Wisconsin dataset for \(\entropy\) (yellow) and \(\globeps\) (blue).
	The green ARC corresponds to the combination of \(\entropy\) and \(\globeps\) with \(\gamma=0.53\).
	The point cloud (logarithmic scale) depicts for each instance if its predicted class was correct (green) or wrong (red).}
	\label{fig:arc_and_cloud}
\end{figure}
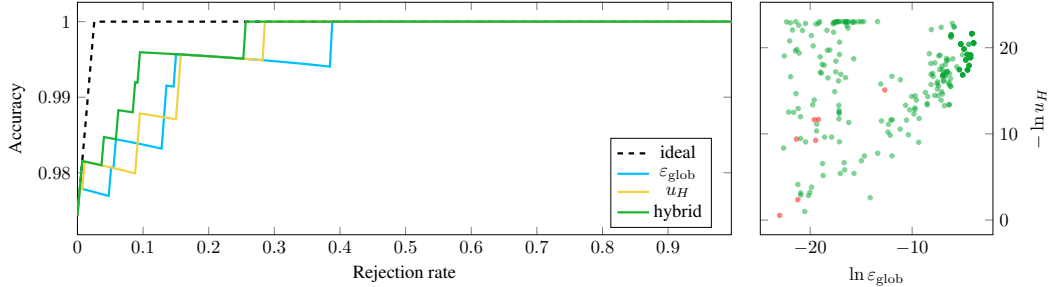

Accuracy rejection curves (ARC) offer a visual way to evaluate this \citep{pmlr-v8-nadeem10a}.
For a given reliability metric, an ARC is made by first ordering all instances in order of increasing reliability; so from high to low uncertainty for uncertainty metrics, or from low to high robustness for robustness metrics.
Once the order is determined, we start rejecting instances in that order, such that the ones with the lowest reliability get rejected first, and at every step we calculate the accuracy of the remaining instances.
So, in essence, ARCs plot the accuracy as a function of the rejection rate.
Figure~\ref{fig:arc_and_cloud} (left side) displays an example of such an ARC for both an uncertainty metric (yellow) and a robustness metric (blue).
Note that the higher the overall curve is for a given metric, the better, with the ideal case being a strictly increasing curve (black).

As can be seen from Figure~\ref{fig:arc_and_cloud}, comparing the ARCs of different reliability metrics is not always straightforward.
In this example, which of the two (yellow or blue) performs best depends on the rejection rate.
This makes assigning a winner a subjective matter. To address this issue, we try to summarize the quality of such a curve in a single value. Following the approach suggested in the conclusion of the work of \citet{pmlr-v8-nadeem10a}, we use the area under the ARC, which we'll refer to as AU-ARC.
Since an ARC consists of a discrete number of points, the AU-ARC can simply be calculated by taking the average of the accuracies that correspond to all possible rejection rates.
For the ARCs in Figure~\ref{fig:arc_and_cloud} of the uncertainty and robustness metric, the AU-ARCS are \(0.9968\) and \(0.9961\), respectively, with uncertainty slightly outperforming robustness here.

To assess the performance of the different uncertainty and robustness metrics we consider, we conducted experiments on the following datasets from the UCI Machine Learning Repository \citep{ucimlrepository}:
Adult, 
Australian Credit Approval, 
Bank Marketing, 
Breast Cancer Wisconsin, 
German Credit Data, 
Lymphography, 
National Poll on Healthy Aging (NPHA), 
Nursery, 
Solar Flare, 
SPECT Heart 
and Student Performance. 
Since we restrict ourselves to discrete features, we decided to remove the continuous ones from the datasets where these were present.
We also chose to remove instances with missing values.
For two particular datasets we adapted the task to one that leans more toward standard classification.
The Solar Flare dataset has originally three possible target variables that can be predicted, being the number of flares that occur, and this for three types of flares;
we turned this into a binary classification task whose aim it is to predict if at least one solar flare occurs (of any type).
Similarly, for the Student Performance dataset we predict whether the student passes or fails, instead of predicting the exact grade.
Unless the dataset provides a test set itself, we randomly split the datasets into \(\trainset\) and \(\testset\), containing 60\% and 40\% respectively.
We used the Naive Bayes Classifier throughout all of our experiments.
To learn this model, we first optimize a smoothing parameter with 5-fold cross validation on \(\trainset\).
Once the optimal smoothing parameter is found, we train the classifier on the entire training set for said parameter.

Figure~\ref{fig:arc_and_cloud} was made for the Breast Cancer Wisconsin dataset, and with \(\entropy\) as uncertainty metric and \(\globeps\) as robustness metric.
The AU-ARCs we mentioned earlier for their ARCs can also be found in Table~\ref{tab:auarc} in the row of `Breast Cancer', in the second and third column.
This table furthermore contains AU-ARCs for all the other datasets.
The yellow column shows the AU-ARC for the considered uncertainty metric, in this case \(\entropy\); the blue columns show the AU-ARC of the robustness metrics \(\globeps\) and \(\loceps\).
Comparing the AU-ARC of the uncertainty and robustness metrics, we see that there is no clear winner throughout all the datasets.
Similar tables for \(\aleatunc\), \(\epistunc\), \(\maxprob\) and \(\totunc\) are available in the \nameref{sec:appendix}, with similar conclusions.

\begin{table}[!ht]
	\footnotesize
	\centering
	\begin{filecontents*}{table_aucs_entropies.csv}
dataset,uncertainty,global,hybridA,mixingA,local,hybridB,mixingB
Adult,0.9295,0.7690,0.9295,1.00,0.9066,0.9295,1.00
Austr. Cr.,0.9236,0.8872,0.9246,0.86,0.9139,0.9265,0.75
Bank M.,0.9485,0.9299,0.9481,0.88,0.9452,0.9485,0.55
BCW,0.9968,0.9961,0.9978,0.53,0.9962,0.9974,0.52
German Cr.,0.8338,0.7972,0.8376,0.85,0.8380,0.8378,0.53
Heart dis.,0.7602,0.6761,0.7600,0.95,0.7540,0.7602,0.95
Lymphogr.,0.9440,0.8981,0.9425,0.88,0.9419,0.9428,0.77
NPHA,0.4962,0.5159,0.4913,0.96,0.5021,0.4917,0.77
Nursery,0.9813,0.9730,0.9814,0.91,0.9822,0.9824,0.28
Solar (big),0.8603,0.8693,0.8836,0.71,0.8926,0.8874,0.23
Solar (small),0.8709,0.7990,0.8797,0.78,0.8597,0.8666,0.19
SPECT,0.9458,0.5738,0.9457,0.99,0.8915,0.9458,0.99
Stud. Math,0.9434,0.9205,0.9445,0.60,0.9465,0.9468,0.31
Stud. Port,0.8898,0.8952,0.9067,0.79,0.9276,0.9093,0.77
\end{filecontents*}
\csvreader[
        myNames,
        tabularray =
            {
                rowsep = 0.5mm,
                colsep = 3mm,
                vline{3,6} = {.6pt, black, dashed},
                hline{1,2} = {.6pt, black},
                row{1} = {c},
				column{1} = {l},
                column{2} = {unc_yellow!25,c},
                column{3,6} = {rob_blue!25,c},
                column{4,7} = {hybrid_green!50,c},
                cell{2}{4} = {font=\bfseries},
                cell{2}{7} = {font=\bfseries},
                cell{3}{4} = {font=\bfseries},
                cell{3}{7} = {font=\bfseries},
                cell{4}{7} = {font=\bfseries},
                cell{5}{4} = {font=\bfseries},
                cell{5}{7} = {font=\bfseries},
                cell{6}{4} = {font=\bfseries},
                cell{7}{7} = {font=\bfseries},
                cell{10}{4} = {font=\bfseries},
                cell{10}{7} = {font=\bfseries},
                cell{11}{4} = {font=\bfseries},
                cell{12}{4} = {font=\bfseries},
                cell{13}{7} = {font=\bfseries},
                cell{14}{4} = {font=\bfseries},
                cell{14}{7} = {font=\bfseries},
                cell{15}{4} = {font=\bfseries},
            },
        table head = { Dataset & \(\entropy\) & \(\varepsilon_{\mathrm{glob}}\) & hybrid & \(\gamma\) & \(\varepsilon_{\mathrm{loc}}\) & hybrid & \(\gamma\) \\ },
    ]{table_aucs_entropies.csv}{}{
            \csvexpval\dataset
        &   \csvexpval\unc
        &   \csvexpval\glob
        &   \csvexpval\hybridA
        &   \csvexpval\mixingA
        &   \csvexpval\local
        &   \csvexpval\hybridB
        &   \csvexpval\mixingB
    }
    \vspace{.6cm}
	\caption{\footnotesize The AU-ARC of \(\entropy\) (yellow), \(\globeps\) (blue), \(\loceps\) (blue) and of the combinations of the robustness metrics with the uncertainty one (green).
	The white columns show the \(\gamma\) used to combine the uncertainty and robustness metrics.}
	\label{tab:auarc}
\end{table}

\section{Combining uncertainty and robustness}
As has become clear in the previous section, it seems that both uncertainty and robustness metrics are capable of assessing the reliability of predictions.
Furthermore, which of them is better seems to depend on the particular dataset.
We now proceed to investigate whether we can combine both metrics to arrive at an even better reliability assessment.
To understand why this might indeed be possible, we take a look at the point cloud on Figure~\ref{fig:arc_and_cloud} (right side).
This point cloud represents each instance of \(\testset\) with a colored dot, where green means that the instance was classified correctly, and red otherwise.
The \(x\)- and \(y\)-coordinate of a dot respectively represent the values for \(\globeps\) and \(\entropy\) on a logarithmic scale.
Since the points are spread over the plane, it means that for a given value of one of the metrics, the other metric could be used to further distinguish the more and less reliable instances.
We also clearly see that misclassified instances tend to have both high uncertainty and low robustness (the red dots on are in the bottom left region).
This indicates that combining the two metrics could lead to an even better reliability assessment.

Given the complementary behavior of uncertainty and robustness metrics, it seems logical to construct a hybrid reliability metric as a function that maps two numerical values, being the metrics we'd like to combine, to a new numerical reliability value that performs even better at ordering instances than either of the two on their own.
It is not obvious, however, how to meaningfully combine two numerical values that capture different concepts of reliability into a single value.
Since we only want to order the instances, we therefore omit the step of constructing a hybrid metric, but instead directly aim to obtain a hybrid order of the instances.

To obtain an order that combines uncertainty and robustness, we take a weighted average of the two orders.
First, we order all instances with both metrics separately to obtain for each instance two numbers that correspond to its position in each of the orders.
If for the \(i\)-th instance the position according to an uncertainty metric $u$ is \(n_{u, i}\), and according to a robustness metric \(\varepsilon\) is \(n_{\varepsilon, i}\), we determine the hybrid position of this instance using the weighted average of the two separate positions. In particular, we let
\[
    h_i \coloneq \gamma n_{u,i} + (1-\gamma)n_{\varepsilon,i},
\]
where the weighting coefficient \(\gamma \in [0,1]\) determines the relative importance of uncertainty and robustness, and then order all instances in order of increasing \(h_i\), where ties are decided by the uncertainty metric.
In particular, \(\gamma=1\) leads to the same order as induced by uncertainty alone, and \(\gamma=0\) to the one for robustness.

Since we've already observed that the relative performance of robustness and uncertainty depends on the dataset, it is clear that the weighting coefficient \(\gamma\) should depend on the dataset.
We therefore choose to optimize \(\gamma\) on the training set.
To do so, we compute the AU-ARC for a grid of possible values for \(\gamma\) and choose the \(\gamma\) that yields the highest AU-ARC for the training set.

For the Breast Cancer Wisconsin dataset, and with \(\entropy\) as uncertainty metric and \(\globeps\) as robustness metric, the result of the hybrid order can be seen in Figure~\ref{fig:arc_and_cloud} (left) as the green ARC.
Here, it is visually clear that the order that combines both metrics clearly outperforms the individual ones, as the hybrid ARC lies above the other two.
The AU-ARC of the three ARCs shown on this figure are \(0.9968\) for \(\entropy\), \(0.9961\) for \(\globeps\) and \(0.9978\) for the combination. Since \(\gamma=0.53\), robustness and uncertainty contributed more or less equally for this dataset.
The results for the other datasets, and with \(\entropy\) as uncertainty metric, are given in Table~\ref{tab:auarc}.
The green columns contain the AU-ARC of the hybrid order obtained by combing the robustness metric (blue) of the column to the left of it with the uncertainty metric of the yellow column.
The white columns provide the trained weighting coefficient \(\gamma\) used for combining the two metrics.
To make the results more easily interpretable, we highlighted the AU-ARC of the hybrid order in bold whenever it was the highest.
In most cases, the combination of uncertainty and robustness wins (indicated in bold) or is a close second; the only exception seems to be the NPHA dataset. Similar results for (combinations with) \(\aleatunc\), \(\epistunc\), \(\maxprob\) and \(\totunc\) are available in the \nameref{sec:appendix}; the conclusions are mostly similar, except for $\epistunc$, where there is no clear winner between the hybrid approach and uncertainty.
We conclude from these experiments that uncertainty quantification and robustness quantification are not only different on a conceptual level, but that also in practice they have their own way of contributing to assessing the reliability of the predictions of a classifier.

In addition to better reliability assessments, our approach of combining uncertainty and robustness furthermore provides us with information about the relative importance of uncertainty and robustness for each dataset, in the form of the trained weighting coefficient $\gamma$. As can be seen from Table~\ref{tab:auarc} and the additional tables in the \nameref{sec:appendix}, this relative importance varies substantially between the datasets, and furthermore depends on the type of uncertainty and robustness that is considered.

\section{Discussion}
The take-away message of this contribution, in our view, is that robustness quantification provides a valuable tool for assessing the reliability of the predictions of a classifier, especially so, if it is combined with uncertainty quantification. There is, however, still much to explore.

A first straightforward extension to our work would be to combine more than two reliability metrics, instead of combining a single uncertainty metric with a single robustness metric.
This could not only lead to even better results, but it could also give more insight in what metrics are useful and which ones are not, for example by studying the learned weighting coefficients.
Another obvious line of future research would be to try to construct a hybrid reliability metric, instead of focussing solely on constructing a hybrid order.

As for how to evaluate our approach, there are also some alternatives we would like to explore.
We now used AU-ARC, which is a simple and intuitive way of evaluating the overall performance of a reliability metric; however it does not say it all.
Looking back at Figure~\ref{fig:arc_and_cloud}, the hybrid approach clearly outperforms uncertainty for small rejection rates: the difference in accuracy gets up to more than 1\% (e.g. for rejection rate \(0.1\)), reducing the percentage of misclassified instances by more than half.
Nevertheless, the difference between the AU-ARCs (\(0.9978\) for the hybrid approach and \(0.9968\) for uncertainty) is almost negligible because the performance of both approaches is identical for higher rejection rates. An AU-ARC thus not entirely captures the performance, at least not if we have a particular rejection rate in mind. For that reason, in our future work, we'd like to extend our approach to the situation where (information about) the rejection rate is known beforehand, and learn $\gamma$ such as to optimize for that setting rather than simply optimize the average accuracy with AU-ARCs.

Finally, we've shown in earlier work that, for synthetic data, robustness quantification outperforms uncertainty quantification in the presence of distribution shift and limited data \citep{detavernier2025robustness}. It would therefore be interesting to study the performance of our hybrid approach in such settings as well, for example by considering real datasets where distribution shift is present, or by artificially reducing the size of our datasets to study its effect on performance.

\section*{Acknowledgements}
We would like to thank the anonymous reviewer for their time, kind words and helpful feedback.
The work of both authors was partially supported by Ghent University's Special Research Fund, through Jasper De Bock's starting grant number 01N04819.

\bibliographystyle{plainnat}
\bibliography{detavernier_eiml25}

\newpage
\section*{Appendix}\label{sec:appendix}
\begin{figure}[!ht]
    \centering
	\begin{tikzpicture}[scale=0.7]
		\begin{axis}[
			xlabel={Rejection rate},
			ylabel={Accuracy},
			width=14cm,
			height=6cm,
			enlarge x limits=false,
			legend pos=south east,
			legend entries={%
                ideal,
				$\globeps$,
				$\aleatunc$,
				hybrid,
			},
			]
			\addplot+ [no marks, very thick, color=black, dashed] table [x={rej_rate}, y={ideal}] {solar_big_global_eps_aleatoric.dat};
			\addplot+ [no marks, very thick, color=rob_blue] table [x={rej_rate}, y={robustness}] {solar_big_global_eps_aleatoric.dat};
			\addplot+ [no marks, very thick, color=unc_yellow] table [x={rej_rate}, y={uncertainty}] {solar_big_global_eps_aleatoric.dat};
			\addplot+ [no marks, very thick, color=hybrid_green] table [x={rej_rate}, y={hybrid}] {solar_big_global_eps_aleatoric.dat};
		\end{axis}
	\end{tikzpicture}
    \caption{\footnotesize ARCs for the Solar Flare (big) dataset for \(\aleatunc\) (yellow), \(\globeps\) (blue), the hybrid order (green, \(\gamma=0.66\)) and the optimal curve (black).}
    \label{fig:aleat}
\end{figure}
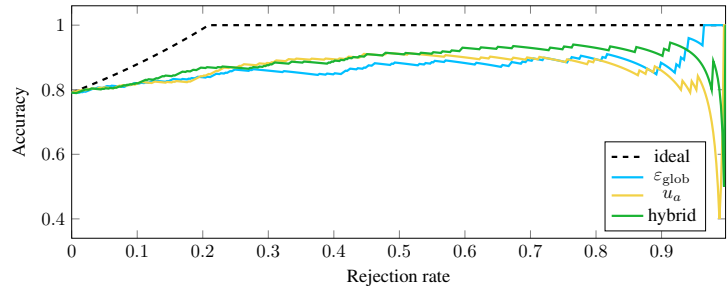

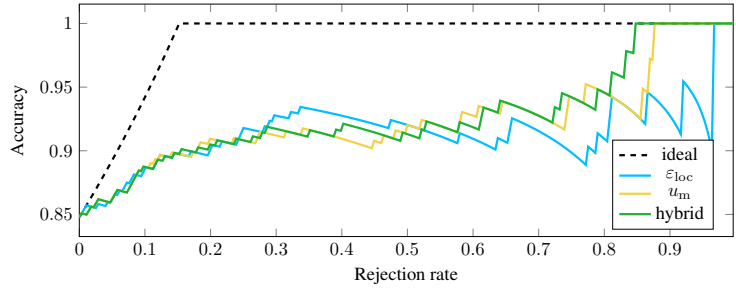
\begin{figure}[!ht]
    \centering
	\begin{tikzpicture}[scale=0.7]
		\begin{axis}[
			xlabel={Rejection rate},
			ylabel={Accuracy},
			width=14cm,
			height=6cm,
			enlarge x limits=false,
			legend pos=south east,
			legend entries={%
                ideal,
				$\loceps$,
				$\maxprob$,
				hybrid,
			},
			]
			\addplot+ [no marks, very thick, color=black, dashed] table [x={rej_rate}, y={ideal}] {Australian_local_eps_max_prob.dat};
			\addplot+ [no marks, very thick, color=rob_blue] table [x={rej_rate}, y={robustness}] {Australian_local_eps_max_prob.dat};
			\addplot+ [no marks, very thick, color=unc_yellow] table [x={rej_rate}, y={uncertainty}] {Australian_local_eps_max_prob.dat};
			\addplot+ [no marks, very thick, color=hybrid_green] table [x={rej_rate}, y={hybrid}] {Australian_local_eps_max_prob.dat};
		\end{axis}
	\end{tikzpicture}
    \caption{\footnotesize ARCs for the Australian Credit dataset for \(\maxprob\) (yellow), \(\loceps\) (blue), the hybrid order (green, \(\gamma=0.75\)) and the optimal curve (black).}
    \label{fig:maxprob}
\end{figure}

To adhere to the page limit constraint, the results in the paper focussed on the uncertainty metric \(\entropy\), comparing it to and combining it with both \(\varepsilon_{\mathrm{glob}}\) and \(\varepsilon_{\mathrm{loc}}\). In this appendix, we present similar results for the uncertainty metrics \(\aleatunc\), \(\epistunc\), \(\maxprob\) and \(\totunc\). The relevant AU-ARCs are available in Tables~\ref{tab:aleat}, \ref{tab:epist}, \ref{tab:maxprob} and \ref{tab:tot}, respectively, similarly to Table~\ref{tab:auarc}. For each uncertainty metric, we also display the ARC of uncertainty, robustness and the hybrid approach, for a handpicked choice of dataset and robustness metric. These ARCs are available in Figures~\ref{fig:aleat}, \ref{fig:epist}, \ref{fig:maxprob} and \ref{fig:tot}, respectively, similarly to the left-hand side of Figure~\ref{fig:arc_and_cloud}.

\subsection*{Results for \(\aleatunc\)}

\begin{table}[!ht]
	\footnotesize
	\centering
	\begin{filecontents*}{table_aucs_aleatoric.csv}
dataset,uncertainty,global,hybridA,mixingA,local,hybridB,mixingB
Adult,0.9298,0.7690,0.9298,1.00,0.9066,0.9298,1.00
Australian Credit,0.9214,0.8872,0.9218,0.84,0.9139,0.9230,0.79
Bank Marketing,0.9483,0.9299,0.9481,0.91,0.9452,0.9487,0.66
Breast Cancer,0.9967,0.9961,0.9978,0.55,0.9962,0.9973,0.50
German Credit,0.8335,0.7972,0.8388,0.78,0.8380,0.8386,0.27
Heart disease,0.7608,0.6761,0.7592,0.94,0.7540,0.7608,0.99
Lymphography,0.9464,0.8981,0.9448,0.68,0.9419,0.9476,0.32
NPHA,0.4900,0.5159,0.4843,0.95,0.5021,0.4849,0.69
Nursery,0.9813,0.9730,0.9813,0.93,0.9822,0.9824,0.17
Solar Flare (big),0.8640,0.8693,0.8887,0.66,0.8926,0.8883,0.18
Solar Flare (small),0.8811,0.7990,0.8812,0.70,0.8597,0.8801,0.47
SPECT Heart,0.9512,0.5738,0.9512,1.00,0.8915,0.9511,0.99
Student Math,0.9501,0.9205,0.9470,0.60,0.9465,0.9492,0.51
Student Port,0.8878,0.8952,0.9039,0.78,0.9276,0.8995,0.86
\end{filecontents*}
\csvreader[
        myNames,
        tabularray =
            {
                rowsep = 0.5mm,
                colsep = 3mm,
                vline{3,6} = {.6pt, black, dashed},
                hline{1,2} = {.6pt, black},
                row{1} = {c},
                column{2} = {unc_yellow!25,c},
                column{3,6} = {rob_blue!25,c},
                column{4,7} = {hybrid_green!50,c},
                cell{2}{4} = {font=\bfseries},
                cell{2}{7} = {font=\bfseries},
                cell{3}{4} = {font=\bfseries},
                cell{3}{7} = {font=\bfseries},
                cell{4}{7} = {font=\bfseries},
                cell{5}{4} = {font=\bfseries},
                cell{5}{7} = {font=\bfseries},
                cell{6}{4} = {font=\bfseries},
                cell{6}{7} = {font=\bfseries},
                cell{7}{7} = {font=\bfseries},
                cell{8}{7} = {font=\bfseries},
                cell{10}{4} = {font=\bfseries},
                cell{10}{7} = {font=\bfseries},
                cell{11}{4} = {font=\bfseries},
                cell{12}{4} = {font=\bfseries},
                cell{13}{4} = {font=\bfseries},
                cell{15}{4} = {font=\bfseries}
            },
        table head = { Dataset & \(\aleatunc\) & \(\varepsilon_{\mathrm{glob}}\) & hybrid & \(\gamma\) & \(\varepsilon_{\mathrm{loc}}\) & hybrid & \(\gamma\) \\ },
    ]{table_aucs_aleatoric.csv}{}{
            \csvexpval\dataset
        &   \csvexpval\unc
        &   \csvexpval\glob
        &   \csvexpval\hybridA
        &   \csvexpval\mixingA
        &   \csvexpval\local
        &   \csvexpval\hybridB
        &   \csvexpval\mixingB
    }
    \vspace{.6cm}
	\caption{\footnotesize The AU-ARC of \(\aleatunc\) (yellow), \(\globeps\) (blue), \(\loceps\) (blue) and of the combinations of the robustness metrics with the uncertainty one (green).
	The white columns show the \(\gamma\) used to combine the uncertainty and robustness metrics.}
    \label{tab:aleat}
\end{table}

\newpage

\subsection*{Results for \(\epistunc\)}

\begin{table}[!ht]
	\footnotesize
	\centering
	\begin{filecontents*}{table_aucs_epistemic.csv}
dataset,uncertainty,global,hybridA,mixingA,local,hybridB,mixingB
Adult,0.9158,0.7690,0.9158,1.00,0.9066,0.9158,1.00
Australian Credit,0.9075,0.8872,0.8981,0.89,0.9139,0.9069,0.87
Bank Marketing,0.9403,0.9299,0.9399,0.95,0.9452,0.9428,0.63
Breast Cancer,0.9967,0.9961,0.9977,0.53,0.9962,0.9973,0.43
German Credit,0.8126,0.7972,0.8172,0.85,0.8380,0.8375,0.06
Heart disease,0.7663,0.6761,0.7631,0.93,0.7540,0.7634,0.62
Lymphography,0.9318,0.8981,0.9294,0.82,0.9419,0.9425,0.07
NPHA,0.4657,0.5159,0.4657,1.00,0.5021,0.4989,0.03
Nursery,0.9732,0.9730,0.9788,0.49,0.9822,0.9828,0.19
Solar Flare (big),0.8353,0.8693,0.8721,0.60,0.8926,0.8926,0.00
Solar Flare (small),0.8755,0.7990,0.8631,0.76,0.8597,0.8697,0.51
SPECT Heart,0.9384,0.5738,0.9384,1.00,0.8915,0.9384,1.00
Student Math,0.9526,0.9205,0.9505,0.79,0.9465,0.9522,0.93
Student Port,0.8835,0.8952,0.8858,0.96,0.9276,0.8879,0.95
\end{filecontents*}
\csvreader[
        myNames,
        tabularray =
            {
                rowsep = 0.5mm,
                colsep = 3mm,
                vline{3,6} = {.6pt, black, dashed},
                hline{1,2} = {.6pt, black},
                row{1} = {c},
                column{2} = {unc_yellow!25,c},
                column{3,6} = {rob_blue!25,c},
                column{4,7} = {hybrid_green!50,c},
                cell{2}{4} = {font=\bfseries},
                cell{2}{7} = {font=\bfseries},
                cell{5}{4} = {font=\bfseries},
                cell{5}{7} = {font=\bfseries},
                cell{6}{4} = {font=\bfseries},
                cell{8}{7} = {font=\bfseries},
                cell{10}{4} = {font=\bfseries},
                cell{10}{7} = {font=\bfseries},
                cell{11}{4} = {font=\bfseries},
                cell{11}{7} = {font=\bfseries},
                cell{13}{4} = {font=\bfseries},
                cell{13}{7} = {font=\bfseries}
            },
        table head = { Dataset & \(\epistunc\) & \(\varepsilon_{\mathrm{glob}}\) & hybrid & \(\gamma\) & \(\varepsilon_{\mathrm{loc}}\) & hybrid & \(\gamma\) \\ },
    ]{table_aucs_epistemic.csv}{}{
            \csvexpval\dataset
        &   \csvexpval\unc
        &   \csvexpval\glob
        &   \csvexpval\hybridA
        &   \csvexpval\mixingA
        &   \csvexpval\local
        &   \csvexpval\hybridB
        &   \csvexpval\mixingB
    }
    \vspace{.6cm}
	\caption{\footnotesize The AU-ARC of \(\epistunc\) (yellow), \(\globeps\) (blue), \(\loceps\) (blue) and of the combinations of the robustness metrics with the uncertainty one (green).
	The white columns show the \(\gamma\) used to combine the uncertainty and robustness metrics.}
    \label{tab:epist}
\end{table}

\begin{figure}[!ht]
    \centering
	\begin{tikzpicture}[scale=0.7]
		\begin{axis}[
			xlabel={Rejection rate},
			ylabel={Accuracy},
			width=14cm,
			height=6cm,
			enlarge x limits=false,
			legend pos=south east,
			legend entries={%
                ideal,
				$\globeps$,
				$\epistunc$,
				hybrid,
			},
			]
			\addplot [no marks, very thick, color=black, dashed] table [x={rej_rate}, y={ideal}] {nursery_global_eps_epistemic.dat};
			\addplot [no marks, very thick, color=rob_blue] table [x={rej_rate}, y={robustness}] {nursery_global_eps_epistemic.dat};
			\addplot+ [no marks, very thick, color=unc_yellow] table [x={rej_rate}, y={uncertainty}] {nursery_global_eps_epistemic.dat};
			\addplot+ [no marks, very thick, color=hybrid_green] table [x={rej_rate}, y={hybrid}] {nursery_global_eps_epistemic.dat};
		\end{axis}
	\end{tikzpicture}
    \caption{\footnotesize ARCs for the Nursery dataset for \(\epistunc\) (yellow), \(\globeps\) (blue), the hybrid order (green, \(\gamma=0.49\)) and the optimal curve (black).}
    \label{fig:epist}
\end{figure}

\newpage

\subsection*{Results for \(\maxprob\)}

\begin{table}[!ht]
	\footnotesize
	\centering
	\begin{filecontents*}{table_aucs_max_prob.csv}
dataset,uncertainty,global,hybridA,mixingA,local,hybridB,mixingB
Adult,0.9295,0.7690,0.9295,1.00,0.9066,0.9295,1.00
Australian Credit,0.9236,0.8872,0.9246,0.86,0.9139,0.9265,0.75
Bank Marketing,0.9485,0.9299,0.9481,0.88,0.9452,0.9485,0.55
Breast Cancer,0.9968,0.9961,0.9978,0.53,0.9962,0.9974,0.52
German Credit,0.8338,0.7972,0.8376,0.85,0.8380,0.8378,0.53
Heart disease,0.7617,0.6761,0.7617,1.00,0.7540,0.7617,0.99
Lymphography,0.9432,0.8981,0.9435,0.88,0.9419,0.9427,0.77
NPHA,0.4986,0.5159,0.4976,0.99,0.5021,0.4986,1.00
Nursery,0.9803,0.9730,0.9803,1.00,0.9822,0.9822,0.00
Solar Flare (big),0.8603,0.8693,0.8836,0.71,0.8926,0.8874,0.23
Solar Flare (small),0.8709,0.7990,0.8797,0.78,0.8597,0.8666,0.19
SPECT Heart,0.9458,0.5738,0.9457,0.99,0.8915,0.9458,0.99
Student Math,0.9434,0.9205,0.9445,0.60,0.9465,0.9468,0.31
Student Port,0.8898,0.8952,0.9067,0.79,0.9276,0.9093,0.77
\end{filecontents*}
\csvreader[
        myNames,
        tabularray =
            {
                rowsep = 0.5mm,
                colsep = 3mm,
                vline{3,6} = {.6pt, black, dashed},
                hline{1,2} = {.6pt, black},
                row{1} = {c},
                column{2} = {unc_yellow!25,c},
                column{3,6} = {rob_blue!25,c},
                column{4,7} = {hybrid_green!50,c},
                cell{2}{4} = {font=\bfseries},
                cell{2}{7} = {font=\bfseries},
                cell{3}{4} = {font=\bfseries},
                cell{3}{7} = {font=\bfseries},
                cell{4}{7} = {font=\bfseries},
                cell{5}{4} = {font=\bfseries},
                cell{5}{7} = {font=\bfseries},
                cell{6}{4} = {font=\bfseries},
                cell{7}{4} = {font=\bfseries},
                cell{7}{7} = {font=\bfseries},
                cell{8}{4} = {font=\bfseries},
                cell{10}{4} = {font=\bfseries},
                cell{10}{7} = {font=\bfseries},
                cell{11}{4} = {font=\bfseries},
                cell{12}{4} = {font=\bfseries},
                cell{13}{7} = {font=\bfseries},
                cell{14}{4} = {font=\bfseries},
                cell{14}{7} = {font=\bfseries},
                cell{15}{4} = {font=\bfseries},
            },
        table head = { Dataset & \(\maxprob\) & \(\varepsilon_{\mathrm{glob}}\) & hybrid & \(\gamma\) & \(\varepsilon_{\mathrm{loc}}\) & hybrid & \(\gamma\) \\ },
    ]{table_aucs_max_prob.csv}{}{
            \csvexpval\dataset
        &   \csvexpval\unc
        &   \csvexpval\glob
        &   \csvexpval\hybridA
        &   \csvexpval\mixingA
        &   \csvexpval\local
        &   \csvexpval\hybridB
        &   \csvexpval\mixingB
    }
    \vspace{.6cm}
	\caption{\footnotesize The AU-ARC of \(\maxprob\) (yellow), \(\globeps\) (blue), \(\loceps\) (blue) and of the combinations of the robustness metrics with the uncertainty one (green).
	The white columns show the \(\gamma\) used to combine the uncertainty and robustness metrics.}
    \label{tab:maxprob}
\end{table}

\newpage

\subsection*{Results for \(\totunc\)}
\begin{table}[!ht]
	\footnotesize
	\centering
	\begin{filecontents*}{table_aucs_total.csv}
dataset,uncertainty,global,hybridA,mixingA,local,hybridB,mixingB
Adult,0.9297,0.7690,0.9297,1.00,0.9066,0.9297,1.00
Australian Credit,0.9222,0.8872,0.9230,0.86,0.9139,0.9246,0.72
Bank Marketing,0.9488,0.9299,0.9487,0.91,0.9452,0.9490,0.65
Breast Cancer,0.9966,0.9961,0.9978,0.55,0.9962,0.9972,0.43
German Credit,0.8330,0.7972,0.8376,0.82,0.8380,0.8367,0.63
Heart disease,0.7651,0.6761,0.7618,0.92,0.7540,0.7651,0.99
Lymphography,0.9433,0.8981,0.9448,0.84,0.9419,0.9456,0.47
NPHA,0.4875,0.5159,0.4794,0.97,0.5021,0.4688,0.80
Nursery,0.9814,0.9730,0.9814,0.92,0.9822,0.9824,0.22
Solar Flare (big),0.8624,0.8693,0.8827,0.75,0.8926,0.8798,0.52
Solar Flare (small),0.8790,0.7990,0.8683,0.69,0.8597,0.8793,0.56
SPECT Heart,0.9489,0.5738,0.9489,1.00,0.8915,0.9489,1.00
Student Math,0.9512,0.9205,0.9482,0.67,0.9465,0.9511,0.91
Student Port,0.8874,0.8952,0.8917,0.91,0.9276,0.8874,1.00
\end{filecontents*}
\csvreader[
        myNames,
        tabularray =
            {
                rowsep = 0.5mm,
                colsep = 3mm,
                vline{3,6} = {.6pt, black, dashed},
                hline{1,2} = {.6pt, black},
                row{1} = {c},
                column{2} = {unc_yellow!25,c},
                column{3,6} = {rob_blue!25,c},
                column{4,7} = {hybrid_green!50,c},
                cell{2}{4} = {font=\bfseries},
                cell{2}{7} = {font=\bfseries},
                cell{3}{4} = {font=\bfseries},
                cell{3}{7} = {font=\bfseries},
                cell{4}{7} = {font=\bfseries},
                cell{5}{4} = {font=\bfseries},
                cell{5}{7} = {font=\bfseries},
                cell{6}{4} = {font=\bfseries},
                cell{7}{7} = {font=\bfseries},
                cell{8}{4} = {font=\bfseries},
                cell{8}{7} = {font=\bfseries},
                cell{10}{4} = {font=\bfseries},
                cell{10}{7} = {font=\bfseries},
                cell{11}{4} = {font=\bfseries},
                cell{12}{7} = {font=\bfseries},
                cell{13}{4} = {font=\bfseries},
                cell{13}{7} = {font=\bfseries},
            },
        table head = { Dataset & \(\totunc\) & \(\varepsilon_{\mathrm{glob}}\) & hybrid & \(\gamma\) & \(\varepsilon_{\mathrm{loc}}\) & hybrid & \(\gamma\) \\ },
    ]{table_aucs_total.csv}{}{
            \csvexpval\dataset
        &   \csvexpval\unc
        &   \csvexpval\glob
        &   \csvexpval\hybridA
        &   \csvexpval\mixingA
        &   \csvexpval\local
        &   \csvexpval\hybridB
        &   \csvexpval\mixingB
    }
    \vspace{.6cm}
	\caption{\footnotesize The AU-ARC of \(\totunc\) (yellow), \(\globeps\) (blue), \(\loceps\) (blue) and of the combinations of the robustness metrics with the uncertainty one (green).
	The white columns show the \(\gamma\) used to combine the uncertainty and robustness metrics.}
    \label{tab:tot}
\end{table}

\begin{figure}[!ht]
    \centering
	\begin{tikzpicture}[scale=0.7]
		\begin{axis}[
			xlabel={Rejection rate},
			ylabel={Accuracy},
			width=14cm,
			height=6cm,
			enlarge x limits=false,
			legend pos=south east,
			legend entries={%
                ideal,
				$\globeps$,
				$\totunc$,
				hybrid,
			},
			]
			\addplot+ [no marks, very thick, color=black, dashed] table [x={rej_rate}, y={ideal}] {german_global_eps_total.dat};
			\addplot+ [no marks, very thick, color=rob_blue] table [x={rej_rate}, y={robustness}] {german_global_eps_total.dat};
			\addplot+ [no marks, very thick, color=unc_yellow] table [x={rej_rate}, y={uncertainty}] {german_global_eps_total.dat};
			\addplot+ [no marks, very thick, color=hybrid_green] table [x={rej_rate}, y={hybrid}] {german_global_eps_total.dat};
		\end{axis}
	\end{tikzpicture}
    \caption{\footnotesize ARCs for the German Credit dataset for \(\totunc\) (yellow), \(\globeps\) (blue), the hybrid order (green, \(\gamma=0.82\)) and the optimal curve (black).}
    \label{fig:tot}
\end{figure}

\end{document}